\documentclass[letterpaper]{article} 
\usepackage{aaai25}  
\usepackage{times}  
\usepackage{helvet}  
\usepackage{courier}  
\usepackage[hyphens]{url}  
\usepackage{graphicx} 
\usepackage{natbib}  
\usepackage{caption} 
\usepackage{algorithm}
\usepackage{algorithmic}
\usepackage{comment}
\usepackage{booktabs}
\usepackage{subcaption}

\usepackage{xcolor}

\usepackage{tcolorbox}
\tcbuselibrary{listingsutf8}
\definecolor{customblue}{RGB}{143,170,220}

\usepackage{newfloat}
\usepackage{listings}
\DeclareCaptionStyle{ruled}{labelfont=normalfont,labelsep=colon,strut=off} 
\floatstyle{ruled}
\newfloat{listing}{tb}{lst}{}
\floatname{listing}{Listing}
\usepackage{amsmath}
\usepackage{amsfonts}
\usepackage{booktabs}
\usepackage{tabularx}

\title{CareLab at \#SMM4H-HeaRD 2025: Insomnia Detection and Food Safety Event Extraction with Domain-Aware Transformers}

\author{
    Zihan Liang\equalcontrib,
    Ziwen Pan\equalcontrib,
    Sumon Kanti Dey,
    Azra Ismail
}
\affiliations{
    Emory University \\
    Atlanta, Georgia, United States\\
{\{zihan.liang, ziwen.pan, sumon.kanti.dey, azra.ismail\}}@emory.edu
}

\usepackage{bibentry}

\begin{document}

\maketitle
\begin{abstract}
This paper presents our system for the SMM4H–HeaRD 2025 shared tasks, specifically Task 4 (Subtasks 1, 2a, and 2b) and Task 5 (Subtasks 1 and 2). Task 4 focused on detecting mentions of insomnia in clinical notes, while Task 5 addressed the extraction of food safety events from news articles. We participated in all subtasks and report key findings across them, with particular emphasis on Task 5 Subtask 1, where our system achieved strong performance---securing first place with an F1 score of 0.958 on the test set. To attain this result, we employed encoder-based models (e.g., RoBERTa), alongside GPT-4 for data augmentation. This paper outlines our approach, including preprocessing, model architecture, and subtask-specific adaptations.
\end{abstract}

\section{Introduction}
The increasing digitization of healthcare and public health communications has led to a proliferation of unstructured text data from clinical records and regulatory announcements. This presents significant opportunities for natural language processing (NLP) to support early risk detection, medical condition monitoring, and health policy response. In this study, we describe our participation in two shared tasks from the SMM4H 2025 initiative, both centered on real-world textual data but addressing distinct domains and challenges. Task 4 involves detecting evidence of insomnia in clinical notes from the MIMIC-III database. Task 5 focuses on identifying food safety-related events---specifically product recalls and disease outbreaks---from news articles and FDA press releases. Together, these tasks highlight the growing need for domain-adapted NLP systems capable of extracting actionable insights from complex health-related text.

Although the two tasks differ in scope, they reflect complementary challenges in applied healthcare NLP. Task 4 requires a nuanced understanding of medical terminology, latent symptom expression, and structured diagnostic guidelines. In contrast, Task 5 emphasizes sentence-level event detection in the public health domain, which benefits from entity-rich expressions and formal news language. Our system achieved top performance in Task 5 Subtask 1, ranking first among all participants with an F1 score of 0.958 in classifying sentences into recall, outbreak, or neither categories. Through careful data augmentation, domain-specific preprocessing, and transformer-based modeling, we demonstrate strong performance in sentence classification.

While our system did not reach leading performance in the span-based subtasks, the comparative results across Task 4 and Task 5 (Subtask 2) highlight how task complexity, label sparsity, and evidence requirements influence NLP model behavior in clinical and public health domains. This study contributes to the growing literature on domain-adaptive NLP and raises further questions about designing robust pipelines for multi-layered textual inference in health-focused tasks.

\section{Datasets}
\subsection{Task 4: MIMIC-III Clinical Database}
The dataset used in this task is a subset of the MIMIC-III v1.4 database \cite{johnson2016mimic}, which contains over two million de-identified clinical notes from ICU admissions. This subset consists of 210 discharge summaries annotated for insomnia using rule-based diagnostic criteria. Each note is labeled with a binary indicator of insomnia presence, five rule-specific diagnostic tags, and corresponding evidence spans. These annotations jointly support three subtasks: overall insomnia classification, multi-label rule-based classification, and extraction of textual evidence, capturing both explicit and implicit expressions of sleep disturbance.

While prior work on sleep disorder detection in MIMIC-III has focused on structured data or general classification \cite{sivarajkumar2022sleep, li2021icu, irving2021psychosis}, this corpus emphasizes explainability through span-level annotation---a direction still underexplored in clinical NLP.

\subsection{Task 5: FORCE Food Safety Corpus}

Task 5 uses the FORCE (Foodborne disease Outbreak and ReCall Event extraction from openweb) dataset \cite{jana2024force}, comprising 8,100 English news articles and FDA press releases annotated at sentence and entity levels. Each sentence is labeled as either a food recall, an outbreak, or neither. The entity-level annotations cover six types of entities: \textbf{Organization (Org)} refers to the responsible company or agency; \textbf{Product (Prdt)} identifies the contaminated or recalled food item; \textbf{Cause (Cau)} captures the underlying incident cause (e.g., contamination, mislabeling); \textbf{Disease (Dis)} denotes the reported illness (e.g., \textit{Listeria}); \textbf{Number of People Affected (\#Aff)} indicates how many individuals were impacted; and \textbf{Location (Loc)} specifies the geographical site of the event. These labels are used in our Subtask 2 evaluation (see Table~\ref{tab:task5_summary}(b)). 

This corpus was annotated by domain expert and underwent quality assurance via inter-annotator agreement to ensure reliability and consistency. Compared to prior datasets \cite{lee2023sfafsm, goel2024recipe}, FORCE provides uniquely aligned sentence-level and entity-level annotations across real-world regulatory documents.


\section{Methodology}
\subsection{Task 4: Modeling Insomnia Detection from Clinical Notes}

We approach the insomnia detection task as a combination of document-level classification and evidence identification, focusing on predicting insomnia status and rule-based indicators from clinical narratives.

To address both Subtask 1 (binary insomnia classification) and Subtask 2A (multi-label rule classification), we implemented and compared three distinct modeling strategies: traditional machine learning baselines, recurrent neural networks, and transformer-based models. Each approach was applied uniformly to both subtasks, with Subtask 2A framed as a multi-output binary classification problem.

First, we trained traditional classifiers---logistic regression, SVM, random forest, and XGBoost—using TF-IDF lexical features. While these models offered the advantage of computational efficiency and provided a solid baseline, they often overfit to surface-level lexical patterns and struggled to capture the deeper contextual meaning of sentences---an essential requirement in the nuanced landscape of healthcare-related text. Their limitations highlighted the need for more context-aware models capable of generalizing beyond keyword cues.

Next, we developed a BiLSTM model with pretrained GloVe embeddings to capture sequential and syntactic dependencies within the text \cite{pennington-etal-2014-glove, dheeraj2021glove}. Using the same model architecture was applied to both subtasks, we applied sigmoid activation to support multi-label predictions. This approach yielded moderate improvements over traditional baselines, particularly in handling syntactic variants of rule expressions.

Finally, we fine-tuned ClinicalBERT---a domain-specific transformer model pretrained on clinical corpora \cite{huang2019clinicalbert}. We used the [CLS] token representation for classification and optimized the model using binary cross-entropy loss across all six labels (one global insomnia label and five rule-specific indicators). ClinicalBERT consistently outperformed other models, benefiting from its contextual embeddings and domain-specific pretraining. The best-performing results are reported in Table~\ref{tab:task4_reformatted}. To address class imbalance, we applied class weighting during neural model training and used SMOTE oversampling for the traditional baselines to boost minority class representation during training.

For Subtask 2B, we developed a rule-based evidence extraction system using curated medical keywords and regular expression patterns. This approach was interpretable and easy to implement. However, it lacked the semantic flexibility needed to handle more complex language. Specifically, it struggled with paraphrased expressions, coreference resolution, and negation, which limited its effectiveness in accurately aligning evidence spans with the annotated text. Recent advancements in span-based pre-training techniques---such as  Span Selection Pre-training \cite{glass2020span}---offer promising improvements by teaching models to select spans from relevant passages rather than relying solely on token-level memorization, which could help address paraphrased or indirect expressions. 

\subsection{Task 5: Food Recall and Outbreak Detection in News}

We approach Task 5 as a combination of sentence-level classification and entity-level information extraction over food safety–related news articles. Subtask 1 involved classifying each sentence into one of three categories: Food Recall, Foodborne Disease Outbreak, or Neither. Subtask 2 required extracting key entities such as product names, infection types, and affected populations.

For Subtask 1, we fine-tune an encoder-based model, RoBERTa \cite{liu2019roberta}, to perform 3-way sentence classification. To mitigate strong class imbalance---especially for the \texttt{Neither} category---we adopt a two-stage balancing strategy. First, we augmented the training dataset with GPT-4-generated examples \cite{achiam2023gpt} that resembled neutral content from FDA documents and improved the classifier's ability to distinguish informative sentences from generic text. An example of the prompt used for this augmentation is shown in Figure~\ref{fig:gpt4_prompt}. Second, we applied class weighting during training to penalize errors on minority classes more heavily. The loss function is defined as:

\[
\mathcal{L}_{\text{weighted}} = - \sum_{c=1}^{C} w_c \cdot \mathbf{1}_{\{y=c\}} \cdot \log \hat{y}_c
\]

where the weight of the class \(w_c\) is inversely proportional to the empirical frequency \(\pi_c\) of the class \(c\). This combination of LLM-based augmentation and class-weighted loss significantly improved recall for underrepresented classes while maintaining high overall accuracy.

To further boost robustness, we used ensemble inference by averaging predictions across five independently trained models using different random seeds. The final system achieved the highest macro-F1 and accuracy scores in Subtask 1 (F1 = 0.958, Acc = 0.959), demonstrating the combined benefit of data-centric augmentation and transformer-based modeling.

For Subtask 2, we adopted a rule-based pipeline that detects entities using spaCy's dependency parser \footnote{https://spacy.io/api/dependencyparser} combined with hand-crafted regular expressions. Each entity type is associated with a matching function that extracts candidate text spans from an input sentence. For example, infection-related terms are extracted as:

\begin{align*}
\mathcal{E}_{\text{infection}}(x) = \{s \in x \mid\  
&\texttt{regex\_match}( \\
& s,\text{r``Listeria|Salmonella|E.coli"}) \}
\end{align*}

Although this approach is interpretable and effective for surface-form patterns, it lacks generalization to semantically implied mentions. As a result, it struggles with indirect expressions and exhibits low recall, particularly in fields such as infection causes and disease names.

\begin{figure}[htbp]
\centering
\begin{tcolorbox}[colback=gray!5!white, colframe=customblue!90!black,
                  boxrule=0.5mm, arc=4mm, auto outer arc,
                  fonttitle=\bfseries, width=\columnwidth,
                  title=Prompt template]

You are an FDA press release writer. Your task is to write a formal, FDA-style press release (250–300 words) announcing an update related to food labeling or packaging. 

\textbf{Instructions:}
\begin{itemize}
    \item Do \textbf{NOT} discuss food recalls, disease outbreaks, or safety incidents.
    \item Do \textbf{NOT} include contact information, phone numbers, emails, or links.
    \item Ensure each press release is unique---avoid duplicating previous wording, structure, or content.
    \item Maintain a professional, public-health tone.
    \item Vary the focus slightly each time (e.g., nutrition facts, allergen labeling, ingredient transparency, packaging design updates).
\end{itemize}

\end{tcolorbox}
\caption{Example prompt used for synthetic data generation.}
\label{fig:gpt4_prompt}
\end{figure}

\section{Results}
In this section, we present the experimental results of our models on the official shared tasks for Tasks 4 and 5. We report the evaluation metrics defined by the organizers for each subtask, including F1-score for binary and multi-label classification, ROUGE scores for evidence extraction, and F1-scores for entity-level information extraction.


\subsection{Task 4 Results: Insomnia Detection}

\begin{table}[ht]
\centering
\resizebox{\linewidth}{!}{%
\begin{tabular}{lcccc}
\toprule
\textbf{Task 4 Subtask 1} & \textbf{Model} & \textbf{F1-score} & \textbf{Precision} & \textbf{Recall} \\
\midrule
CareLab   & ClinicalBERT & 0.786 & 0.648 & 1.000 \\
\textit{Mean}   & -- & 0.877 & 0.853 & 0.913 \\
\textit{Median} & -- & 0.869 & 0.840 & 0.935 \\
\midrule
\textbf{Task 4 Subtask 2A} & \textbf{Model} & \textbf{F1-score} & \textbf{Precision} & \textbf{Recall} \\
\midrule
CareLab   & ClinicalBERT & 0.769 & 0.758 & 0.780 \\
\textit{Mean}   & -- & 0.718 & 0.673 & 0.789 \\
\textit{Median} & -- & 0.692 & 0.650 & 0.819 \\
\midrule
\textbf{Task 4 Subtask 2B} & \textbf{Model} & \textbf{F1-score} & \textbf{Precision} & \textbf{Recall} \\
\midrule
CareLab   & Regex Rule-based & 0.135 & 0.097 & 0.372 \\
\textit{Mean}   & -- & 0.386 & 0.423 & 0.443 \\
\textit{Median} & -- & 0.446 & 0.514 & 0.487 \\
\bottomrule
\end{tabular}
}
\caption{Task 4 evaluation summary showing the performance of our best models (ClinicalBERT and rule-based extractor) across subtasks, compared with task-level benchmarks.}
\label{tab:task4_reformatted}
\end{table}

Table~\ref{tab:task4_reformatted} presents our Task 4 results alongside task-wide mean and median scores. For Subtask 1, ClinicalBERT achieved perfect recall (1.000) but relatively low precision (0.648), resulting in an F1-score of 0.786—below the task mean (0.877) and median (0.869). This suggests a tendency to overpredict positives, likely due to class imbalance and few explicit negative indicators.


For Subtask 2A (rule-level multi-label classification), ClinicalBERT outperformed task-level baselines with an F1-score of 0.769 (precision 0.758, recall 0.780). These results highlight ClinicalBERT's effectiveness in capturing multiple structured diagnostic criteria within clinical narratives. To further mitigate label imbalance, we applied SMOTE-based augmentation, which improved the model's ability to recognize underrepresented rule-specific labels.

In contrast, our approach to Subtask 2B relied on a hand-crafted, regex-based span extractor built on insomnia-related keyword dictionaries. The regex-based extractor, while interpretable and efficient, yielded low performance (F1 = 0.135; precision = 0.098; recall = 0.372). Its rule-dependence made it brittle to paraphrasing, negation, and complex syntax, underscoring the value of context-aware, span-based transformers for robust evidence extraction.


\subsection{Task 5 Results: Food Recall and Outbreak Detection}

\begin{table}[ht]
\begin{minipage}{\linewidth}
\centering
{\small \textbf{(a) Subtask 1: Sentence-level classification}}
\label{tab:task5_subtask1}
\resizebox{0.8\linewidth}{!}{
\begin{tabular}{@{}llcccc@{}}
\toprule
\textbf{Team} & \textbf{Model} & \textbf{F1-score} & \textbf{Precision} & \textbf{Recall} & \textbf{Accuracy} \\
\midrule
CareLab & RoBERTa + Data & \textbf{0.958} & \textbf{0.957} & \textbf{0.959} & \textbf{0.959} \\
 & Augmentation &  & &  &  \\
Best    & --                           & \textbf{0.958} & \textbf{0.957} & \textbf{0.959} & \textbf{0.959} \\
\bottomrule
\end{tabular}
}
\end{minipage}

\vspace{1em}

\begin{minipage}{\linewidth}
{\small \textbf{(b) Subtask 2: Entity-level extraction}}
\centering
\resizebox{\linewidth}{!}{
\begin{tabular}{@{}p{1.3cm}p{1.8cm}ccccccc@{}}
\toprule
\textbf{Team} & \textbf{Model} & \textbf{Avg.} & \textbf{Org} & \textbf{Prdt} & \textbf{Cau} & \textbf{Dis} & \textbf{\#Aff} & \textbf{Loc} \\
\midrule
CareLab & Rule-based & 0.119 & 0.120 & 0.140 & 0.010 & 0.000 & 0.390 & 0.230 \\
& Extractor & & & & & & & \\
Best & -- & \textbf{0.576} & 0.940 & 0.620 & 0.240 & 0.640 & 0.700 & 0.600 \\
\bottomrule
\end{tabular}
}
\end{minipage}
\caption{Performance comparison for Task 5 subtasks. (a) Sentence classification and (b) Entity-level extraction.}
\label{tab:task5_summary}
\end{table}


As shown in Table~\ref{tab:task5_summary}(a), our system achieved top performance in Task 5 Subtask 1, with a macro-F1 score of \textbf{0.958}, precision of \textbf{0.957}, and recall of \textbf{0.959}---matching the best-performing team. We used a RoBERTa-large model with aggressive data augmentation to improve generalization, particularly for the underrepresented \texttt{Neither} class. This reflects three key design choices: (1) FDA-derived augmentation to increase class diversity (2) applying class-weighted focal loss to handle severe class imbalance; and (3) incorporating feature-enhanced modeling and uncertainty-aware post-processing to boost classification robustness.

In contrast, Task 5 Subtask 2 required entity-level information extraction. We adopted a Our rule-based extractor performed substantially worse than the top system (F1: 0.119 vs. 0.576), with weak results across most entity types, as shown in Table~\ref{tab:task5_summary}(b). particularly for \texttt{Cause and Disease}, where recall was nearly zero. Our pipeline relied on regular expressions and handcrafted heuristics to identify six types of entities from food safety news, including product names, infection types, and target organizations.

While the rule-based approach worked reasonably for surface-form mentions---such as explicit brand names or known infections like \textit{Listeria}---it failed to generalize to paraphrased, implicit, or nested entities due to its limited coverage and lack of contextual reasoning, leading to particularly low recall and a large performance gap compared to sequence-labeling methods.

\section{Error Analysis}
\subsection{Task 4}
In subtask 1, which involved binary insomnia classification, the ClinicalBERT model demonstrated high recall but relatively lower precision in predicting insomnia presence, suggesting a tendency to over-predict positive cases. This is likely due to the scarcity of explicitly negative examples in the training data and the subtle linguistic cues associated with the absence of insomnia. In many instances, notes contain ambiguous or borderline mentions that were interpreted as positive by the model, especially when no strong negation was present. For example, discharge summaries often list medications like trazodone or melatonin "as needed at bedtime," without context suggesting active insomnia symptoms. In other cases, mentions of depression or blunted affect triggered false positives despite no reference to sleep difficulties. Such cases highlight the limitations of token-level cues in the absence of discourse-level understanding. Although class weighting helped to some extent, future work could explore contrastive pretraining or hard negative mining to reduce false positives.

In Subtask 2A, ClinicalBERT achieved strong overall performance on the rule-level multi-label classification task. However, the complexity of individual labels varied. Rules involving explicit terminology (e.g., mentions of ``sleep latency $>$ 30 mins") were easier for the model to learn. In contrast, rules that required temporal reasoning or inference---such as identifying chronic patterns---introduced greater label noise and reduced performance. Additionally, some discharge notes also lacked complete documentation, which led to missing labels and reduced learning signals. Incorporating section-specific encoding and temporal context modeling may help improve per-rule classification consistency.

The rule-based system in Subtask 2B showed low performance (F1 = 0.135), revealing several deeper limitations beyond surface-form mismatch. First, clinical notes often describe insomnia in indirect, nuanced ways—using idiomatic expressions, subjective complaints, or embedded temporal cues—that rule-based systems are ill-equipped to capture. Second, the span annotations are often not contiguous or syntactically isolated, making it difficult for regex patterns to delineate precise evidence boundaries. Third, the system lacked the ability to reason over sentence-level discourse structure, which is crucial when the evidence is scattered across multiple clauses or requires linking antecedents and consequences. These challenges were further compounded by class imbalance and limited training data, which led to an over-reliance on high-precision but low-recall rules, ultimately reducing the system's coverage and generalizability.

\subsection{Task 5}
In subtask 1, the RoBERTa-based classifier achieved high performance in the sentence-level classification task, but some sources of error persisted. Misclassifications were most common between the ``Outbreak" and ``Recall" classes, especially in cases where overlapping or incomplete information appeared in the same sentence. In some instances, the classifier assigned the correct label only after data augmentation with GPT-4-generated negative examples, highlighting issues related to class imbalance and lexical similarity in the original training data.

In Subtask 2, our rule-based entity extraction system achieved limited performance (F1 = 0.119), falling significantly behind the top-performing system (F1 = 0.576). This performance gap is primarily due to the system's heavy reliance on surface-form pattern matching and its inability to incorporate contextual understanding. A major limitation was poor recall, especially for implicit or semantically diverse expressions. For example, the system frequently failed to identify generic references such as ``the item" or ``the contamination," which require contextual cues to resolve. It also struggled with paraphrased expressions of key entities---for instance, when the text described the cause of an infection in an indirect way (e.g., ``due to improper handling" instead of explicitly stating ``bacterial contamination"). Similarly, abstract or implied mentions of affected populations (such as ``those impacted" or ``the group at risk") were often overlooked, as the system lacked the semantic reasoning needed to associate these phrases with specific entity types.

Moreover, many gold entities were embedded in complex or non-canonical sentence structures, such as passive voice, nested clauses, or ellipses, which the regular expression engine failed to capture. The rule-based approach also struggled with coreference resolution and cross-type disambiguation, leading to frequent confusion when the same phrase could indicate multiple entity types depending on context. These limitations highlight the system’s lack of generalization and robustness in handling real-world, linguistically diverse news texts.

To address these challenges, future work will investigate context-aware neural sequence labeling models, such as CRF-augmented transformers trained on the FORCE corpus. These models can leverage rich contextual cues and capture entity boundaries more flexibly. Joint span-sequence frameworks, such as those proposed by Nguyen et al.~\cite{nguyen2022jointly}, may further improve extraction by modeling both sparse entity spans and their global document context. We also plan to explore weak supervision and distant supervision techniques to expand coverage for low-resource entity types.

\section{Conclusion}


This paper presents our approach to SMM4H-HeaRD 2025 shared tasks. In Task 4, ClinicalBERT and BiLSTM models achieved strong results in rule-based classification (Subtask 2A) and competitive performance in binary classification (Subtask 1), while a rule-based span extractor (Subtask 2B) underperformed due to limited flexibility in capturing indirect insomnia mentions.

In Task 5, a RoBERTa-based classifier with LLM-driven data augmentation and class balancing delivered top-ranked results in sentence-level event classification (Subtask 1). However, our rule-based entity extractor in Subtask 2 demonstrated limited generalizability, highlighting the importance of supervised, span-aware architectures in real-world public health applications.


Overall, our findings underscore the value of domain-specific pretraining, data augmentation, and architectural alignment with task structure. In the future, we aim to explore span-aware architectures---like QA-style finetuning and contrastive learning \cite{zheng2024enhancing}---to improve both factual grounding and extraction precision. Our ultimate goal is to develop systems that not only perform well but also better reflect the complexity and richness of health-related language in real-world care. Our final codebase for all subtasks is available at: https://github.com/zihanliang/CARELAB-SMM4H2025.


\bibliography{aaai25}

\begin{thebibliography}{15}
\providecommand{\natexlab}[1]{#1}

\bibitem[{Achiam et~al.(2023)Achiam, Adler, Agarwal, Ahmad, Akkaya, Aleman, Almeida, Altenschmidt, Altman, Anadkat et~al.}]{achiam2023gpt}
Achiam, J.; Adler, S.; Agarwal, S.; Ahmad, L.; Akkaya, I.; Aleman, F.~L.; Almeida, D.; Altenschmidt, J.; Altman, S.; Anadkat, S.; et~al. 2023.
\newblock Gpt-4 technical report.
\newblock \emph{arXiv preprint arXiv:2303.08774}.

\bibitem[{Dheeraj and Ramakrishnudu(2021)}]{dheeraj2021glove}
Dheeraj, K.; and Ramakrishnudu, T. 2021.
\newblock Negative emotions detection on online mental-health related patients texts using the deep learning with MHA-BCNN model.
\newblock \emph{Expert Systems with Applications}, 182: 115265.

\bibitem[{Glass et~al.(2020)Glass, Gliozzo, Chakravarti, Ferritto, Pan, Bhargav, Garg, and Sil}]{glass2020span}
Glass, M.; Gliozzo, A.; Chakravarti, R.; Ferritto, A.; Pan, L.; Bhargav, G. P.~S.; Garg, D.; and Sil, A. 2020.
\newblock Span Selection Pre-training for Question Answering.
\newblock In \emph{Proceedings of the 58th Annual Meeting of the Association for Computational Linguistics}, 2773--2782.

\bibitem[{Goel et~al.(2024)Goel, Agarwal, Agrawal, Kapuriya, Konam, Gupta, Rastogi, Niharika, and Bagler}]{goel2024recipe}
Goel, M.; Agarwal, A.; Agrawal, S.; Kapuriya, J.; Konam, A.~V.; Gupta, R.; Rastogi, S.; Niharika; and Bagler, G. 2024.
\newblock Deep Learning Based Named Entity Recognition Models for Recipes.
\newblock \emph{arXiv preprint arXiv:2402.17447}.
\newblock Available at \url{https://arxiv.org/abs/2402.17447}.

\bibitem[{Huang, Altosaar, and Ranganath(2019)}]{huang2019clinicalbert}
Huang, K.; Altosaar, J.; and Ranganath, R. 2019.
\newblock Clinicalbert: Modeling clinical notes and predicting hospital readmission.
\newblock \emph{arXiv preprint arXiv:1904.05342}.

\bibitem[{Irving and et~al.(2021)}]{irving2021psychosis}
Irving, J.; and et~al. 2021.
\newblock Using NLP on Electronic Health Records to Enhance Detection and Prediction of Psychosis Risk.
\newblock \emph{Journal of Biomedical Informatics}, 117: 103772.

\bibitem[{Jana, Sinha, and Dasgupta(2024)}]{jana2024force}
Jana, S.; Sinha, M.; and Dasgupta, T. 2024.
\newblock FORCE: A Benchmark Dataset for Foodborne Disease Outbreak and Recall Event Extraction from News.
\newblock \emph{Proceedings of the 9th SMM4H Workshop}.

\bibitem[{Johnson et~al.(2016)Johnson, Pollard, Shen et~al.}]{johnson2016mimic}
Johnson, A.~E.; Pollard, T.~J.; Shen, L.; et~al. 2016.
\newblock MIMIC-III, a freely accessible critical care database.
\newblock \emph{Scientific data}, 3: 160035.

\bibitem[{Lee et~al.(2023)Lee, Vasanthakumar, Chen et~al.}]{lee2023sfafsm}
Lee, N.; Vasanthakumar, U.; Chen, R.; et~al. 2023.
\newblock Predictive Food Safety Risk Monitoring using AI Technologies.
\newblock In \emph{IEEE International Conference on ICT for Smart Society (ICISS)}.

\bibitem[{Li, Liu, and et~al.(2021)}]{li2021icu}
Li, H.; Liu, J.; and et~al. 2021.
\newblock Development and Validation of a Clinical Prediction Model for Sleep Disorders in the ICU: A Retrospective Cohort Study.
\newblock \emph{Frontiers in Neuroscience}, 15: 644845.

\bibitem[{Liu et~al.(2019)Liu, Ott, Goyal, Du, Joshi, Chen, Levy, Lewis, Zettlemoyer, and Stoyanov}]{liu2019roberta}
Liu, Y.; Ott, M.; Goyal, N.; Du, J.; Joshi, M.; Chen, D.; Levy, O.; Lewis, M.; Zettlemoyer, L.; and Stoyanov, V. 2019.
\newblock Roberta: A robustly optimized bert pretraining approach.
\newblock \emph{arXiv preprint arXiv:1907.11692}.

\bibitem[{Nguyen et~al.(2022)Nguyen, Vu, Nguyen, and Nguyen}]{nguyen2022jointly}
Nguyen, H.~S.; Vu, H.~M.; Nguyen, T.-A.~D.; and Nguyen, M.-T. 2022.
\newblock Jointly Learning Span Extraction and Sequence Labeling for Information Extraction from Business Documents.
\newblock \emph{arXiv preprint arXiv:2205.13434}.

\bibitem[{Pennington, Socher, and Manning(2014)}]{pennington-etal-2014-glove}
Pennington, J.; Socher, R.; and Manning, C. 2014.
\newblock {G}lo{V}e: Global Vectors for Word Representation.
\newblock In Moschitti, A.; Pang, B.; and Daelemans, W., eds., \emph{Proceedings of the 2014 Conference on Empirical Methods in Natural Language Processing ({EMNLP})}, 1532--1543. Doha, Qatar: Association for Computational Linguistics.

\bibitem[{Sivarajkumar and et~al.(2022)}]{sivarajkumar2022sleep}
Sivarajkumar, H.; and et~al. 2022.
\newblock Extraction of Sleep Information from Clinical Notes of Patients with Alzheimer's Disease Using Natural Language Processing.
\newblock \emph{arXiv preprint arXiv:2204.09601}.

\bibitem[{Zheng et~al.(2024)Zheng, Hui, Liu, and Hirschberg}]{zheng2024enhancing}
Zheng, L.~A.; Hui, Z.; Liu, Z.; and Hirschberg, J. 2024.
\newblock Enhancing Pre-Trained Generative Language Models with Question-Attended Span Extraction on Machine Reading Comprehension.
\newblock In \emph{Proceedings of the 2024 Conference on Empirical Methods in Natural Language Processing}, 10046--10063.

\end{thebibliography}

\end{document}